%% file: iclr2027_conference.tex
\documentclass{article} 
\usepackage{iclr2027_conference,times}

\input{math_commands.tex}

\usepackage{hyperref}
\usepackage{url}
\usepackage{array,booktabs}
\usepackage{graphicx}
\usepackage{capt-of}
\usepackage{wrapfig}
\usepackage{float,needspace}
\usepackage{fvextra}
\usepackage[breakable]{tcolorbox}

\NewDocumentCommand{\zhenwen}
{ mO{} }{\textcolor{purple}{\textsuperscript{\textit{Zhenwen}}\textsf{\textbf{\small[#1]}}}}

\NewDocumentCommand{\yanyu}
{ mO{} }{\textcolor{yellow}{\textsuperscript{\textit{Yanyu}}\textsf{\textbf{\small[#1]}}}}

\NewDocumentCommand{\shanyong}
{ mO{} }{\textcolor{blue}{\textsuperscript{\textit{Shanyong}}\textsf{\textbf{\small[#1]}}}}

\title{Momentum-Coupled Rubric Adaptation for Detailed Image Captioning}

\author{Zhenwen Ji$^{1,2}$, Lei Jin$^1$, Shanyong Wang$^1$, Jiaming Lu$^1$, Chengqiang Lu$^1$, Yi Wu$^1$, \\
\textbf{Yao Hu$^1$, Lizhen Cui$^2$, Yanyu Xu$^2$} \\
$^1$Xiaohongshu Inc. \\
$^2$the Joint SDU-NTU Centre for Artificial Intelligence Research (C-FAIR), Shandong University \\
\texttt{jizhenwen@xiaohongshu.com, xu\_yanyu@sdu.edu.cn} \\
}
\iclrfinalcopy 
\begin{document}

\maketitle

\begin{abstract}
Detailed image captioning requires accurate and comprehensive descriptions of fine-grained visual content, yet caption quality spans factual accuracy, information coverage, and clarity. Compared with conventional methods that rely mainly on high-quality supervision or holistic rewards, rubric-based reinforcement learning decomposes these requirements into explicit criteria and provides targeted, structured feedback. However, existing methods often use separate models for caption generation, rubric construction, and judging, which may lead to inconsistent interpretations across roles. Some dynamic rubric methods alternate updates between the caption policy and rubric generator while keeping the judge fixed, but staged optimization may still leave rubric construction and judging out of step with policy optimization.
We propose MoCo Rubric, a two-stage framework that coordinates these roles. First, role-conditioned, shared-parameter multi-task supervised fine-tuning equips a single vision--language model to serve as the Caption Policy, Rubric Generator, and Rubric Judge. Then, the Generator constructs rubrics online from captions sampled by the current Policy, reference captions, and image evidence. The Judge provides rubric-based rewards, and only the Policy receives GRPO updates. As Policy updates change the candidates being evaluated, we use an exponential moving average of the Policy parameters to update one momentum model shared by the Generator and Judge. This gradual transfer lets both rubric roles track Policy updates without separate RL optimization while smoothing parameter changes that could disrupt their rubric capabilities under direct synchronization. Under bounded Policy updates, we derive an upper bound on the parameter gap between the Policy and momentum model and analyze the smoothing effect of momentum updates. Across five captioning benchmarks, MoCo Rubric achieves an average pairwise win rate of 72.83\%, the best mean rank in blind ranking, and the highest average score in caption-based question answering. Our code is available at \url{https://github.com/a23wen/MoCo-Rubric}.

\end{abstract}

\input{sections/introduction}

\input{sections/related_work}
\input{sections/methodology}
\input{sections/experiments}
\input{sections/conclusion}

\subsection*{AI use statement}
AI systems, including LLMs, did not contribute to the development of the research ideas or the writing of this paper to a degree that would warrant authorship or contributor status. In this work, LLMs were used for open-ended task evaluation and served as objects of study.

\subsection*{Ethics statement}
We study image captioning using publicly available datasets, and our work does not involve private or sensitive data. The proposed training procedure is designed as a general-purpose optimization method. Within the scope of this study, we have not identified specific ethical risks related to fairness, bias, discrimination, privacy, or security. This research was conducted in accordance with established standards of research integrity.

\subsection*{Reproducibility statement}
We provide training details, implementation settings for our method, and evaluation procedures in the main text and appendix to support reproduction of the reported results.

\bibliography{iclr2027_conference}
\bibliographystyle{iclr2027_conference}

\clearpage
\appendix
\section{Implementation and Evaluation Details}
\label{app:implementation-evaluation}

\subsection{Prompt Templates}
\label{app:prompts}
\input{sections/appendix_prompts}

\subsection{Reproduction of RubiCap and EvoLM}
\label{app:baseline-reproduction}

\subsubsection{RubiCap}
\label{app:rubicap-reproduction}
We adapt RubiCap~\citep{huang2026rubicap} to our captioning setting and initialize the Caption Policy from the same shared-SFT checkpoint used by our method.
Before RL training, Gemini 3.1 Pro constructs image-specific rubrics using reference captions generated by Gemini 3.1 Pro and GPT-5.6-terra.
These rubrics are generated once and reused throughout training.
For each sampled caption, GPT-5.6-terra serves as the Rubric Judge, checking whether it satisfies each criterion and aggregating the weighted binary judgments into a scalar reward.
The Policy is then optimized with GRPO, while the Judge remains fixed and the rubrics are neither regenerated nor updated.
This preserves RubiCap's offline rubric-guided optimization while adapting its initialization and evaluation models to our experimental setting.

\subsubsection{EvoLM}
\label{app:evolm-reproduction}
We adapt EvoLM's~\citep{li2026evolmselfevolvinglanguagemodels} alternating Policy--Generator optimization and self-constructed preference pairs to our captioning setting.
The Policy, Generator, and Judge are initialized from the same shared-SFT checkpoint, with the Judge kept fixed throughout training.
Rubric generation uses the same prompt and inputs as our method.
We alternate between 50 steps of Policy optimization and 50 steps of Generator optimization.
For Generator training, preference pairs consist of captions from the current Policy and a checkpoint 50 Policy-update steps earlier, treating the current captions as preferred.
Following EvoLM, the Generator reward combines the Judge-score difference between preferred and dispreferred captions with a format-validity reward, weighted by 0.7 and 0.3, respectively.
Both optimization phases use GRPO, with 600 total updates across the two roles.

\subsection{Caption-Based Question Answering Protocol}
\label{app:qa-protocol}
We follow the decoupled VQA protocol of Prism~\citep{NEURIPS2024_cac9e747}, as adopted in CapRL~\citep{xing2026caprl}, to evaluate the utility of generated captions for downstream question answering.
Each captioning model first describes the input image without access to the associated questions, using the caption generation prompt in Appendix~\ref{app:prompts}.
A fixed text-only language model then receives the generated description and a question, together with answer options when applicable.
The description serves as its sole source of visual information.

We evaluate this pipeline on CaptionQA, BLINK, TextVQA, DocVQA, and ChartQA.
The answering model is held fixed across captioning methods, and its predictions are scored against the reference answers using the corresponding evaluation procedure for each benchmark.
This setting measures how effectively captions preserve information needed for downstream questions under a common answering model.

\section{Supplementary Evaluation Results}
\label{app:supplementary-evaluation}


\subsection{Evaluation with GPT-5.6}
\label{app:gpt56-evaluation}
We repeat the pairwise caption comparison in Table~\ref{tab:pairwise-caption-comparison} with GPT-5.6 Sol (reasoning effort: none) as the evaluator in place of Gemini 3.1 Pro. We keep the generated captions, image--caption pairs, comparison prompt, and Qwen3-VL-8B-Instruct baseline fixed, and use the same win-rate metric. In every comparison, Caption A is the evaluated method's output and Caption B is the Qwen3-VL-8B-Instruct baseline output. Table~\ref{tab:gpt56-pairwise-caption-comparison} reports the results.

\begin{table}[htbp]
\centering
\caption{Pairwise win rates (\%) against Qwen3-VL-8B with GPT-5.6 Sol as the evaluator. Average denotes the benchmark mean. Best results are bold; second-best results are underlined.}
\label{tab:gpt56-pairwise-caption-comparison}
\small
\setlength{\tabcolsep}{4pt}
\begin{tabular*}{\textwidth}{@{\extracolsep{\fill}}lcccccc@{}}
\toprule
Method & PixMo-Cap & DenseFusion & CapArena & CompreCap & DOCCI & Average \\
\midrule
\multicolumn{7}{@{}l}{\textit{Captioning Baselines}} \\
\addlinespace[2pt]
ShareGPT4V-7B & 1.00 & 0.20 & 0.50 & 1.25 & 1.00 & 0.79 \\
RICO-Flash-7B & 5.02 & 4.02 & 5.34 & 8.06 & 4.20 & 5.33 \\
OmniCaptioner-8B & 5.62 & 3.41 & 6.34 & 7.89 & 9.80 & 6.61 \\
JoyCaption-8B & 11.65 & 6.43 & 11.52 & 31.18 & 13.60 & 14.87 \\
MetaCaptioner-8B & 14.06 & 10.64 & 13.52 & 20.97 & 14.80 & 14.80 \\
CapRL-InternVL-8B & 19.88 & 18.67 & 14.52 & 23.12 & 14.80 & 18.20 \\
CapRL-Qwen3VL-4B & 29.12 & 31.33 & 24.04 & 28.32 & 28.80 & 28.32 \\
Qwen3-VL-32B & 52.61 & 50.60 & 49.58 & 52.33 & 48.20 & 50.67 \\
\midrule
\multicolumn{7}{@{}l}{\textit{Rubric-based Methods}} \\
\addlinespace[2pt]
RubiCap & \underline{60.44} & \underline{60.64} & \underline{62.77} & \underline{62.37} & \underline{64.60} & \underline{62.16} \\
EvoLM & 59.84 & 54.82 & 59.27 & 58.06 & 60.40 & 58.48 \\
\midrule
\multicolumn{7}{@{}l}{\textit{Our Framework}} \\
\addlinespace[2pt]
SFT (shared) & 56.83 & 55.42 & 57.76 & 54.84 & 60.20 & 57.01 \\
Ours & \textbf{67.47} & \textbf{67.67} & \textbf{69.45} & \textbf{69.18} & \textbf{70.20} & \textbf{68.79} \\
\bottomrule
\end{tabular*}
\end{table}

MoCo Rubric ranks first on all five benchmarks under GPT-5.6 Sol, with an average win rate of 68.79\%. It exceeds RubiCap by 6.63 percentage points, EvoLM by 10.31 points, and Qwen3-VL-32B by 18.12 points on average. RubiCap ranks second on each benchmark, while both RubiCap and EvoLM outperform the strongest captioning baseline, Qwen3-VL-32B, on average.

Although the absolute win rates vary with the evaluator, MoCo Rubric has the highest average under both judges (72.83\% with Gemini 3.1 Pro and 68.79\% with GPT-5.6 Sol). This consistency supports the compatibility of the pairwise evaluation protocol with different LLM evaluators and shows that our method's relative advantage persists across the two judges. Each benchmark result uses images with valid judgments for all 12 methods (498, 498, 599, 558, and 500 images, respectively); 67 of 31,920 attempted comparisons were excluded by the evaluator's content filter.

\subsection{Human Annotation Agreement}
\label{app:human-agreement}
\input{sections/appendix_human_agreement}

\subsection{Training Reward Dynamics}
\label{app:training-reward}
\begin{figure}[H]
    \centering
    \includegraphics[width=\linewidth]{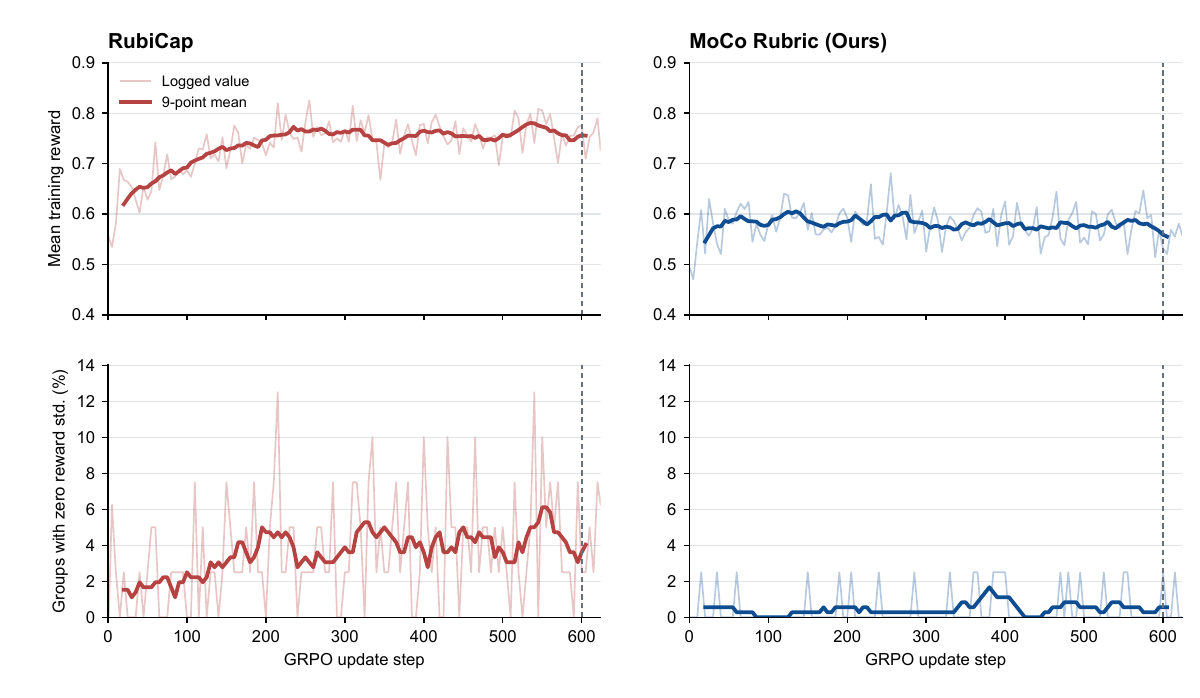}
    \caption{GRPO training dynamics of RubiCap and MoCo Rubric. The upper panels show mean training reward; the lower panels show the fraction of rollout groups with zero within-group reward standard deviation. Light traces are logged values and dark traces are centered nine-point moving averages. Dashed lines mark the evaluated step-600 checkpoints.}
    \label{fig:grpo-training-dynamics}
\end{figure}

Figure~\ref{fig:grpo-training-dynamics} compares the logged mean training reward and \texttt{train/frac\_reward\_zero\_std} for RubiCap and MoCo Rubric. The latter is the fraction of rollout groups whose sampled captions all receive the same reward, yielding no within-group reward advantage for GRPO. Both runs contain 126 logged points over 624 update steps; step 600 is the checkpoint used for external evaluation.

RubiCap's mean reward rises from 0.652 over steps 1--100 to 0.763 over steps 201--300, then remains near 0.758 over steps 301--600. MoCo Rubric does not show the same sustained rise toward a high-reward plateau: its mean reward is 0.588 over steps 201--300 and 0.577 over steps 301--600. Across logged points from steps 5--600, the mean fraction of zero-standard-deviation groups is 3.59\% for RubiCap and 0.46\% for MoCo Rubric; over steps 301--600, it is 4.33\% and 0.58\%, respectively. Thus, fewer rollout groups lack a within-group reward advantage under our method. These are single-run training diagnostics. The two methods use different rubrics and judges, and these runs start from different SFT checkpoints; reward magnitudes are therefore not directly comparable measures of caption quality, and the curves alone do not isolate the cause of the difference.



\clearpage
\section{Qualitative Case Studies}
\label{app:case-studies}

\input{sections/appendix_case_studies}

\section{Proofs and Additional Momentum Analysis}
\label{app}

\subsection{Proof of Proposition 1}
\label{app:tracking-proof}

At synchronization boundary $s$, the momentum update is
\begin{equation}
\bar{\phi}_s
=
m\bar{\phi}_{s-1}
+(1-m)\bar{\theta}_s.
\label{eq:app-ema}
\end{equation}
Define
\begin{equation}
d_s=\bar{\theta}_s-\bar{\phi}_s,
\qquad
\xi_s=\bar{\theta}_s-\bar{\theta}_{s-1},
\end{equation}
so that $e_s=|d_s|$ and $\Delta_s=|\xi_s|$. Substituting Eq.~(\ref{eq:app-ema}) into $d_s$ gives
\begin{align}
d_s
&=
\bar{\theta}_s
-
\left[
m\bar{\phi}_{s-1}
+(1-m)\bar{\theta}_s
\right]
\nonumber\\
&=
m(\bar{\theta}_s-\bar{\phi}_{s-1})
\nonumber\\
&=
m(\xi_s+d_{s-1}).
\label{eq:app-error-recurrence}
\end{align}
Taking norms and applying the triangle inequality yields
\begin{equation}
e_s
\leq
m e_{s-1}+m\Delta_s.
\label{eq:app-one-step-bound}
\end{equation}
Recursively expanding this inequality gives
\begin{equation}
e_s
\leq
m^s e_0
+
\sum_{j=1}^{s}
m^{s-j+1}\Delta_j.
\label{eq:app-expanded-bound}
\end{equation}

Because synchronization occurs every $H$ Policy updates and each update satisfies
$|\theta_{t+1}-\theta_t|\leq\delta$, we have
\begin{align}
\Delta_s
&=
|\bar{\theta}_s-\bar{\theta}_{s-1}|
\nonumber\\
&\leq
\sum_{h=1}^{H}
\left|
\theta_{(s-1)H+h}
-
\theta_{(s-1)H+h-1}
\right|
\nonumber\\
&\leq
H\delta.
\label{eq:app-boundary-drift}
\end{align}
Substituting this result into Eq.~(\ref{eq:app-expanded-bound}) and evaluating the geometric series yields
\begin{align}
e_s
&\leq
m^s e_0
+
H\delta
\sum_{j=1}^{s}m^{s-j+1}
\nonumber\\
&=
m^s e_0
+
\frac{m(1-m^s)}{1-m}H\delta.
\label{eq:app-final-tracking-bound}
\end{align}
Since shared initialization gives
$\bar{\theta}_0=\bar{\phi}_0$ and therefore $e_0=0$,
\begin{equation}
e_s
\leq
\frac{m(1-m^s)}{1-m}H\delta
\leq
\frac{mH\delta}{1-m}.
\label{eq:app-uniform-tracking-bound}
\end{equation}

For the momentum-update magnitude, Eq.~(\ref{eq:app-ema}) gives
\begin{align}
\bar{\phi}_s-\bar{\phi}_{s-1}
&=
(1-m)
(\bar{\theta}_s-\bar{\phi}_{s-1})
\nonumber\\
&=
(1-m)(\xi_s+d_{s-1}).
\label{eq:app-momentum-difference}
\end{align}
Taking norms gives
\begin{equation}
u_s
=
|\bar{\phi}_s-\bar{\phi}_{s-1}|
\leq
(1-m)(e_{s-1}+\Delta_s),
\label{eq:app-update-magnitude}
\end{equation}
which completes the proof. \hfill$\square$

\subsection{Local Drift--Noise Analysis}
\label{app:drift-noise}

We next derive the stochastic smoothing result used in the main text. Consider the local Policy-displacement model
\begin{equation}
\xi_s
=
\bar{\theta}_s-\bar{\theta}_{s-1}
=
Hv+\varepsilon_s,
\label{eq:app-drift-noise}
\end{equation}
where
\begin{equation}
\mathbb{E}[\varepsilon_s]=0,
\qquad
\operatorname{Cov}(\varepsilon_s)=H\Sigma.
\label{eq:app-noise-statistics}
\end{equation}
We assume that $v$ is approximately constant over the local analysis window and that $\varepsilon_s$ is independent across non-overlapping synchronization intervals. Because $d_{s-1}$ depends only on noise from preceding intervals, this assumption also implies that $\varepsilon_s$ is independent of $d_{s-1}$.

Substituting Eq.~(\ref{eq:app-drift-noise}) into Eq.~(\ref{eq:app-error-recurrence}) gives
\begin{equation}
d_s
=
m d_{s-1}
+mHv
+m\varepsilon_s.
\label{eq:app-stochastic-recurrence}
\end{equation}
Under shared initialization, $d_0=0$. Recursively expanding Eq.~(\ref{eq:app-stochastic-recurrence}) yields
\begin{equation}
d_s
=
mH\sum_{j=1}^{s}m^{s-j}v
+
m\sum_{j=1}^{s}m^{s-j}\varepsilon_j.
\label{eq:app-stochastic-expanded}
\end{equation}
Its finite-step mean is
\begin{equation}
\mathbb{E}[d_s]
=
\frac{mH(1-m^s)}{1-m}v.
\label{eq:app-finite-error-mean}
\end{equation}
Using the independence of the noise terms, its covariance is
\begin{align}
\operatorname{Cov}(d_s)
&=
m^2
\sum_{j=1}^{s}
m^{2(s-j)}
\operatorname{Cov}(\varepsilon_j)
\nonumber\\
&=
\frac{
m^2H(1-m^{2s})
}{
1-m^2
}\Sigma.
\label{eq:app-finite-error-covariance}
\end{align}
For $0\leq m<1$, taking $s\rightarrow\infty$ gives
\begin{equation}
\mathbb{E}[d_s]
\longrightarrow
\frac{mH}{1-m}v,
\qquad
\operatorname{Cov}(d_s)
\longrightarrow
\frac{m^2H}{1-m^2}\Sigma.
\label{eq:app-stationary-error}
\end{equation}
Thus, the stationary expected squared tracking error is
\begin{equation}
\mathbb{E}\!\left[|d_s|^2\right]
=
\frac{m^2H^2}{(1-m)^2}|v|^2
+
\frac{m^2H}{1-m^2}
\operatorname{tr}(\Sigma).
\label{eq:app-squared-tracking-error}
\end{equation}

We now consider the momentum-model update
\begin{equation}
q_s
=
\bar{\phi}_s-\bar{\phi}_{s-1}.
\end{equation}
From Eq.~(\ref{eq:app-momentum-difference}),
\begin{equation}
q_s
=
(1-m)
\left(
d_{s-1}+Hv+\varepsilon_s
\right).
\label{eq:app-q-expression}
\end{equation}
Its finite-step mean is
\begin{equation}
\mathbb{E}[q_s]
=
H(1-m^s)v.
\label{eq:app-finite-q-mean}
\end{equation}
Since $d_{s-1}$ and $\varepsilon_s$ are independent,
\begin{align}
\operatorname{Cov}(q_s)
&=
(1-m)^2
\left[
\operatorname{Cov}(d_{s-1})
+H\Sigma
\right]
\nonumber\\
&=
\frac{1-m}{1+m}
(1-m^{2s})H\Sigma.
\label{eq:app-finite-q-covariance}
\end{align}
Taking the stationary limit gives
\begin{equation}
\mathbb{E}[q_s]
\longrightarrow
Hv,
\qquad
\operatorname{Cov}(q_s)
\longrightarrow
\frac{1-m}{1+m}H\Sigma.
\label{eq:app-stationary-q}
\end{equation}

For periodic full copying, $m=0$, so $q_s=\xi_s$ and
$\operatorname{Cov}(q_s)=H\Sigma$. EMA therefore reduces the stationary update covariance, relative to full copying at the same $H$, by the factor
\begin{equation}
\frac{1-m}{1+m}.
\end{equation}
However, Eq.~(\ref{eq:app-stationary-error}) shows that increasing $m$ also increases the mean tracking lag and tracking variance. Similarly, increasing $H$ increases both the systematic displacement and accumulated noise between synchronization boundaries. These results establish a parameter-space tracking--stability trade-off; they do not imply that Policy optimization necessarily improves the task-level quality of rubric generation or judging.

\end{document}

%% file: math_commands.tex
\usepackage{amsmath,amsfonts,bm}

\def\eqref#1{equation~\ref{#1}}

\def\1{\bm{1}}

\DeclareMathAlphabet{\mathsfit}{\encodingdefault}{\sfdefault}{m}{sl}
\SetMathAlphabet{\mathsfit}{bold}{\encodingdefault}{\sfdefault}{bx}{n}



%% file: sections/introduction.tex
\section{Introduction}
\label{sec:introduction}


Detailed image captioning requires accurate and comprehensive descriptions of fine-grained visual content, including objects, attributes, actions, and relationships. 
It has various applications, such as providing rich supervision for vision--language aligning and training~\citep{chen2024sharegpt4v}, helping blind people understand their captured surroundings ~\citep{gurari2020captioning}, and enabling language models to answer visual questions only using descriptions ~\citep{xing2025scalecap}. 
However, considering the fact that one image can admit many valid descriptions and limited reference captions cannot cover every relevant details, one of central challenges is how to involve more reliable feedback on factual accuracy and coverage of important visual information beyond limited references.

\begin{figure}[t]
    \centering
    \includegraphics[width=0.98\linewidth]{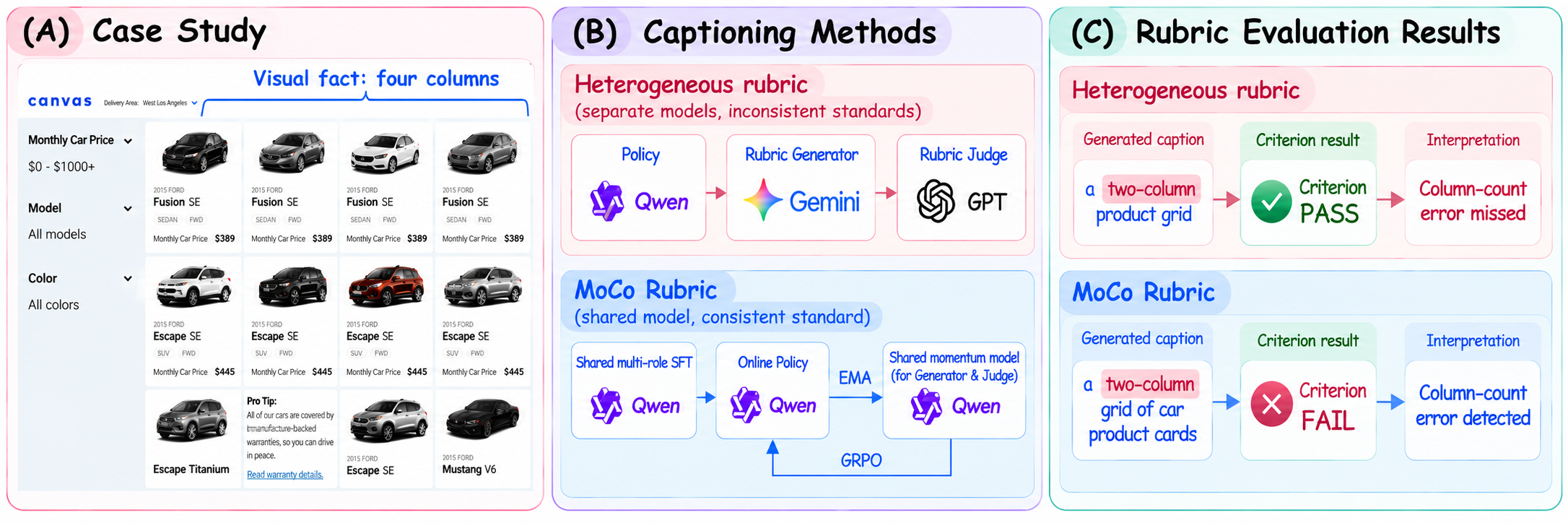}
    \caption{
      Comparison of heterogeneous and shared model distributions in rubric-based caption optimization.
    }
    \label{fig:intro}
\end{figure}

Conventional image captioning methods often rely on high-quality supervised data~\citep{chen2024sharegpt4v,lei2026metacaptioner} or holistic rewards based on caption scores and downstream question-answering utility~\citep{rennie2017self,xing2026caprl} to provide criterion-level yet limited feedback on factual errors and missing visual information.
Very recent Rubric-based reinforcement learning methods instead involve additional feedback by decomposing caption quality into explicit, image-specific criteria and criterion-level judgments into structured rewards~\citep{huang2026rubicap}. 
The rewards are formed through a three-role feedback loop, where a Caption Policy produces candidates, a Rubric Generator defines the criteria, and a Rubric Judge applies them. 
Yet the differences from different backbones and optimization trajectories may lead to semantic inconsistency and missing temporal coordination within the feedback loop.
Firstly, the semantic inconsistency usually arises from the implementation of three roles instantiated as different backbones ~\citep{huang2026rubicap,wang2026coevolvingllmevaluatorspolicies}, such as Rubric Generator using Gemini and Rubric Judge using GPT in Fig. \ref{fig:intro} (a).
For example, Fig.~\ref{fig:intro}(a) shows four columns of car listings, while a candidate caption describes a two-column grid. The Generator defines a broad criterion that checks only whether the caption mentions a grid of cars, and the Judge marks it as satisfied, leaving the column-count error undetected.
Recently, EvoLM ~\citep{wang2026coevolvingllmevaluatorspolicies,li2026evolmselfevolvinglanguagemodels} alternately optimized the Caption Policy and Rubric Generator using same or different backbones to ease such semantic inconsistency, yet the fixed Rubric Judge leads to another challenge of missing temporal coordination. 
The Rubric Generator might overlook important quality differences among the Policy's current candidates, and the Rubric Judge may interpret or apply the resulting criteria differently than intended.
Thus, there occurs a nature question that how rubric construction and judging remain semantically consistent and temporally coordinated with an evolving Caption Policy?

Targeted it, we propose \textsc{MoCo Rubric}, a momentum-coupled rubric adaptation framework for detailed image captioning. 
It includes a Shared Multi-Role SFT stage to semantically align the three roles and a Policy RL with Momentum Transfer stage only optimizing the Caption Policy and gradually transferring its updates to two rubric roles.
Specifically, the Shared Multi-Role SFT targets semantic consistency by combining a common backbone with explicit role-specific supervision. 
Assigning all three roles to a common backbone provides a natural starting point for reducing cross-model semantic variation by placing them in the same visual--language representation space. 
Sharing a backbone alone, however, does not establish the distinct capabilities required to generate captions, formulate evaluation criteria, and apply those criteria reliably.
We therefore employ role-conditioned multi-task supervised fine-tuning on a single vision--language model, explicitly supervising caption generation, rubric construction, and rubric judging. 
The resulting checkpoint initializes both the online Caption Policy and the momentum model shared by the Rubric Generator and Rubric Judge, ensuring that all three roles enter reinforcement learning from the same learned foundation.

Further, we optimize only the Caption Policy with online rubric rewards while updating the model shared by the Rubric Generator and Rubric Judge through momentum transfer, allowing rubric feedback to adapt to the evolving Policy.
Temporal coordination requires the generated rubric content responsive to predicted candidates and the Rubric Generator and Rubric Judge coordinated in parameter space within the same iteration.
Existing dynamic rubric methods such as DynamicRubric and EvoLM address content adaptation by optimizing the Policy and Generator in alternating stages while keeping the Judge fixed~\citep{wang2026coevolvingllmevaluatorspolicies,li2026evolmselfevolvinglanguagemodels}. Such stage-wise alternating optimization refreshes the Generator only at update boundaries, allowing it to lag behind the evolving Policy within each stage, while the Judge remains outside the adaptation process. 
Thus, our MoCo rubrics instead combines candidate-conditioned rubric construction with joint momentum transfer to the Generator and Judge, addressing both content responsiveness and evaluator-state coordination without separate reinforcement learning for either rubric role.
During training, current Policy candidates, reference captions, and image evidence guide rubric construction, the Judge scores the candidates against the shared rubric, and only the Policy is optimized through GRPO~\citep{shao2024deepseekmathpushinglimitsmathematical}.
Because Stage I places all three roles in a shared parameter space, Policy updates may contain changes to visual and linguistic representations relevant across the three tasks. These updates nevertheless optimize caption generation rather than evaluator quality, so directly copying the latest Policy parameters could disrupt the learned rubric capabilities. 
We therefore use an exponential moving average, inspired by MoCo~\citep{he2020momentumcontrastunsupervisedvisual}, to transfer Policy changes gradually to the momentum model shared by the Generator and Judge. 
Both roles use the same lagged parameter version, which remains fixed throughout rubric construction and scoring for each candidate group, thereby coordinating criterion formulation and execution without separate optimization.
We further analyze how the momentum coefficient and synchronization interval jointly control parameter tracking and update smoothness.

Using Qwen3-VL-8B-Instruct~\citep{bai2025qwen3vltechnicalreport} for all three roles, MoCo Rubric achieves a 72.83\% average pairwise win rate across five detailed image captioning benchmarks. It outperforms RubiCap~\citep{huang2026rubicap} by 2.90 percentage points, even though RubiCap employs proprietary models as its Rubric Generator and Rubric Judge. MoCo Rubric also obtains the best mean rank in blind ranking and the highest average score in caption-based question answering, demonstrating improvements in both overall caption quality and downstream utility. 

In summary, we introduce \textsc{MoCo Rubric}, a momentum-coupled rubric adaptation framework for detailed image captioning. 
Shared Multi-Role SFT stage uses a common backbone to establish a shared semantic foundation for the Caption Policy, Rubric Generator, and Rubric Judge. 
Then, we develop Policy RL with Momentum Transfer, which combines candidate-conditioned rubric construction with joint exponential-moving-average transfer of Policy updates to both rubric roles while optimizing only the Caption Policy. Further we analyze how the momentum coefficient and synchronization interval affect parameter tracking and update smoothness. 
Finally, we validate MoCo Rubric on five detailed image captioning benchmarks through pairwise caption comparison, blind ranking, and caption-based question answering, with ablations supporting the complementary effects of Shared Multi-Role SFT, candidate-conditioned rubrics, and joint momentum transfer.


%% file: sections/related_work.tex
\section{Related Work}
\label{sec:related_work}

\textbf{Image Captioning and Evaluation.}
Captioner training combines improved supervision with direct optimization of generation quality. Synthetic captioning, data filtering, and model collaboration provide rich descriptions for supervised learning~\citep{li2022blipbootstrappinglanguageimagepretraining,chen2024sharegpt4v,singla2024pixelsproselargedataset,lei2026metacaptioner}. 
RL has progressed from optimizing sequence-level metrics~\citep{rennie2017self} to rewards tailored to detailed descriptions. Recent methods assess question-answering utility~\citep{xing2026caprl,yang2026caprlunifiedreinforcementlearning}, balance factual correctness, coverage, and linguistic quality~\citep{tang2026cccaptiondualrewardreinforcementlearning,ye2026balcaprlbalancedframeworkrlbased}, or express image-specific requirements through rubrics~\citep{huang2026rubicap}. The approaches extend training beyond imitation of reference captions, making the reward's ability to distinguish evolving policy outputs important. Caption evaluation has expanded from reference matching to semantic and preference-based assessment. BLEU and CIDEr measure agreement with reference text~\citep{papineni-etal-2002-bleu,vedantam2015ciderconsensusbasedimagedescription}, while semantic and cross-modal metrics capture content correspondence~\citep{anderson2016spicesemanticpropositionalimage,hessel-etal-2021-clipscore}. Fine-grained measures examine hallucinations and visual details that aggregate similarity can overlook~\citep{rohrbach-etal-2018-object,dong2024benchmarkingimprovingimagecaption}. For increasingly open-ended descriptions, LLM and VLM judges offer broader quality assessments~\citep{chan-etal-2023-clair,lee-etal-2024-fleur}. CapArena aggregates anonymous human pairwise preferences into model rankings, and its automated counterpart uses VLM comparisons~\citep{cheng2025caparenabenchmarkinganalyzingdetailed}. These complementary tools assess caption quality externally; we focus on adapting structured rewards during training.

\textbf{Rubric-Based Reinforcement Learning.} 
Rubric-based RL decomposes open-ended quality requirements into explicit criteria and aggregates their judgments into rewards~\citep{gunjal2025rubricsrewardsreinforcementlearning,huang2025reinforcementlearningrubricanchors,wang2026meet,viswanathan2025checklistsbetterrewardmodels,yu2025sotopia}. Beyond deriving criteria from instructions and references, rubric construction uses response contrasts, high-quality examples, and recursive refinement to improve specificity and discrimination~\citep{liu2026openrubricsscalablesyntheticrubric,zhang2026chasingtaileffectiverubricbased,shen2026rethinkingrubricgenerationimproving, ye2026auto}. Online approaches condition evaluation criteria on current responses, external evidence, or previously collected rubrics, allowing evaluation content to adapt as the policy changes~\citep{jia2026openrubricsystemscaling,shao2026drtulureinforcementlearning,guan2026evorubricselfevolvingrubricdrivenrl,wang2026coevolvingllmevaluatorspolicies}. Building on explicit, adaptive criteria, we ask whether the parameters that generate and apply them can be driven by Policy RL alone, and examine the reliability and cost of doing so. Training rubric roles extends adaptation from evaluation content to evaluation capability. Rubric-ARM alternates generator and judge optimization~\citep{xu2026alternatingreinforcementlearningrubricbased}, whereas EvoLM and DynamicRubric alternate policy and generator updates with a fixed judge or verifier~\citep{li2026evolmselfevolvinglanguagemodels,wang2026coevolvingllmevaluatorspolicies}. Other approaches jointly optimize response and rubric generation through shared models or separate role adapters~\citep{sheng2026reinforcingchainofthoughtreasoningselfevolving,guan2026evorubricselfevolvingrubricdrivenrl,ding2026evorubricsdynamicrubricsrewards, chen2025decisionflow,yu2026ctm}. These methods introduce learning objectives for the rubric roles to improve evaluation alongside generation. Our approach starts from shared multi-role initialization, applies RL gradients only to the Policy, and updates one momentum model shared by both rubric roles through an exponential moving average of Policy parameters. This parameter-transfer mechanism removes separate RL updates and associated supervision construction for the rubric roles.

%% file: sections/methodology.tex
\section{Methodology}
\label{sec:methodology}

\subsection{Framework Overview}
\label{sec:framework}

\begin{figure}[t]
    \centering
    \includegraphics[width=0.92\linewidth]{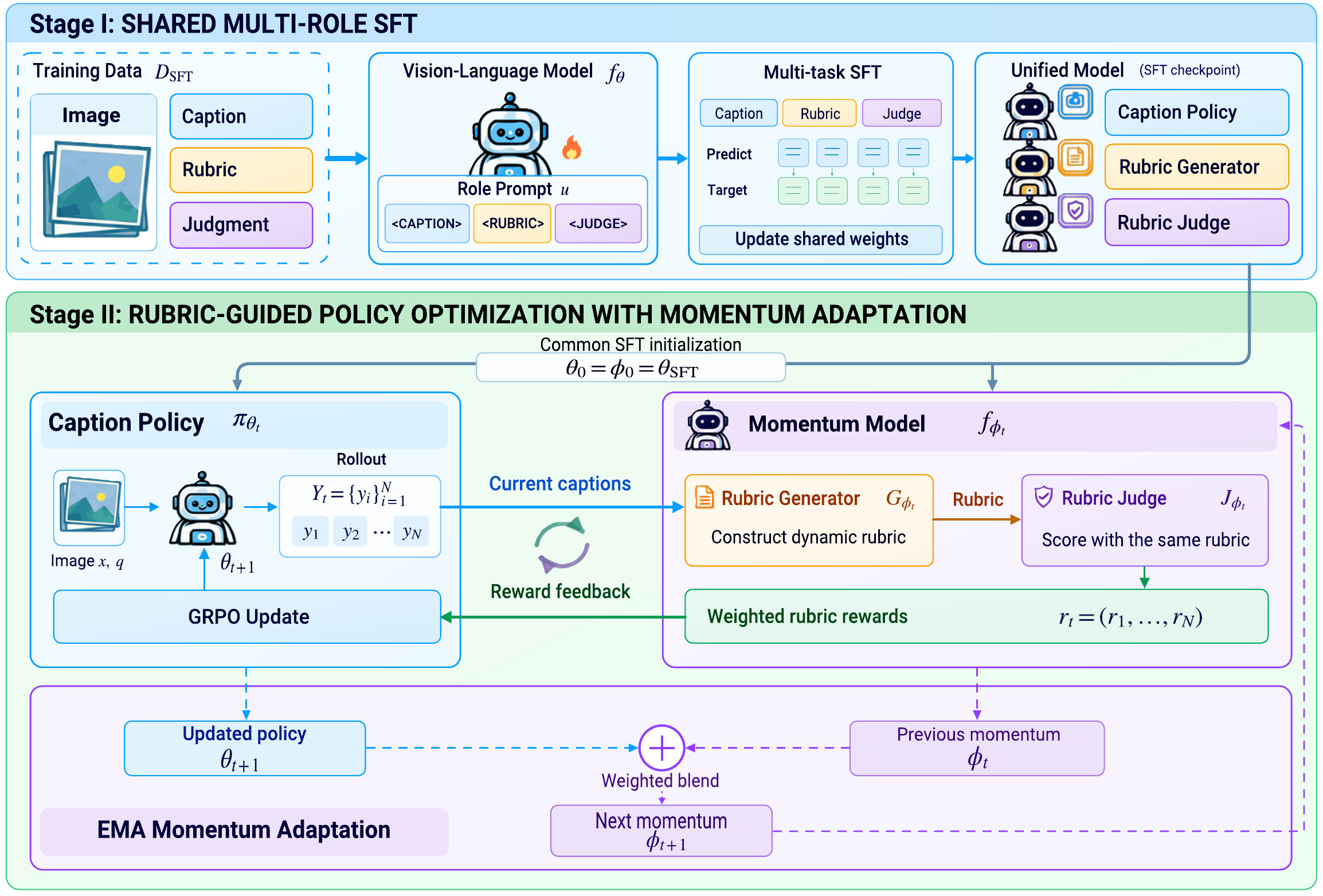}
    \caption{
    Overview of MoCo Rubric. Stage I trains the three roles through shared multi-role SFT. In Stage II, current Policy candidates inform the rubric, the momentum Generator and Judge assign rewards, and only the Policy receives RL gradients. The shared rubric model is updated through EMA at synchronization boundaries.}
    \label{fig:framework}
\end{figure}



Given an image $x$ and a captioning instruction $q$, detailed image captioning aims to generate a description $y$ that is factually accurate and covers important visual information. Because an image can admit multiple valid descriptions, a limited set of reference captions may not capture every relevant detail. We therefore train with image-specific rubric feedback that evaluates generated captions at the criterion level. We denote the Caption Policy by $\pi_\theta$, the Rubric Generator by $G_\phi$, and the Rubric Judge by $J_\phi$; the latter two are the rubric roles. Reference captions are used for rubric construction during training, whereas inference uses only $x$ and $q$ and runs the Caption Policy alone.



This feedback depends on the Policy describing the image, the Generator using it to construct rubrics, and the Judge applying those rubrics to captions. Separately instantiated roles can disagree about caption quality, and Policy updates can change the candidates that the rubric roles need to evaluate. As illustrated in Fig.~\ref{fig:framework}, MoCo Rubric addresses these challenges in two stages. \emph{Shared Multi-Role SFT} explicitly teaches one vision--language model to generate captions, construct rubrics, and judge captions under different role prompts. \emph{Policy RL with Momentum Transfer} then uses current Policy candidates to construct rubrics and optimize only the Policy; an exponential moving average (EMA) transfers its parameter updates to a single momentum model shared by the Generator and Judge. These two stages establish a common role foundation and keep rubric feedback responsive as the Policy evolves.


\subsection{Shared Multi-Role SFT}
\label{sec:parameter-sharing}

Using the same backbone for all three roles provides a common starting point, but does not by itself teach the model how to formulate or apply evaluation criteria. We therefore train one vision--language model $f_\theta$ with role-conditioned, sample-aligned supervision. For a given image, the Policy record pairs $(x,q)$ with a target caption; the Generator record pairs the image, reference captions, and candidate captions with a target rubric; and the Judge records pair a candidate caption and rubric items with criterion-level decisions and rationales. The records are linked by their underlying image and caption instances, while role prompts distinguish the three tasks.



Let $D_{\mathrm{SFT}}$ be the resulting mixed dataset. All three roles update the same parameters through the autoregressive objective
\begin{equation}
    \mathcal{L}_{\mathrm{SFT}}(\theta)
    =\mathbb{E}_{(u,v)\sim D_{\mathrm{SFT}}}
    \left[-\sum_{\ell=1}^{|v|}
    \log p_\theta(v_\ell\mid u,v_{<\ell})\right],
    \label{eq:multi-role-sft}
\end{equation}
where $u$ comprises a role prompt and its task inputs, and $v$ is the corresponding supervision target. We initialize both the online Policy and the momentum model from the resulting checkpoint, $\theta_0=\phi_0=\theta_{\mathrm{SFT}}$. During Stage II, the Generator and Judge use the same parameters $\phi$ under their respective role prompts; only the Policy parameters $\theta$ receive RL gradients. This initialization establishes the rubric-role capabilities before parameter transfer begins.



\subsection{Rubric-Guided Policy Optimization with Momentum Adaptation}


Stage II adapts the content of each rubric to current Policy candidates and transfers Policy parameter changes gradually to the two pretrained rubric roles. 

\textbf{Online rubric construction.} At Policy update $t$, we sample $N=16$ captions from the behavior Policy, $Y_t=\{y_i\}_{i=1}^N$, and select a random subset $S_t\subseteq Y_t$ of $M=4$ captions for rubric construction. Given the reference captions $C^{\mathrm{ref}}$, the Generator produces $R_t=G_{\phi_t}(x,C^{\mathrm{ref}},S_t) =\{(c_k,e_k,w_k)\}_{k=1}^{K_t}$, 
where $c_k$ is a criterion, $e_k$ specifies how to decide whether it is satisfied, and $w_k>0$ is its weight. The subset $S_t$ exposes differences among current candidates, the references supply a quality anchor beyond that subset, and the image provides the factual basis for each criterion. The resulting rubric is applied to every caption in $Y_t$, including captions outside $S_t$.

The Judge receives each caption and the image-grounded rubric, without direct access to the image, and produces a decision $b_{ik}\in\{0,1\}$ for each item, along with a rationale. A value of one means that caption $y_i$ satisfies item $k$. The criterion-level decisions yield a normalized weighted reward $b_{ik}=\bigl[J_{\phi_t}(y_i,R_t)\bigr]_k$, $r_i=\frac{\sum_{k=1}^{K_t}w_k b_{ik}}      {\sum_{k=1}^{K_t}w_k}\in[0,1].$
Using the same rubric and momentum checkpoint for the whole group makes reward differences comparable across candidates. The Judge's rationales explain the item decisions, while only the binary decisions and weights contribute to $r_i$.



\textbf{Policy optimization and momentum transfer.}
We standardize the rewards within each group to obtain GRPO advantages, $A_i=\frac{r_i-\operatorname{mean}_j(r_j)}{\operatorname{std}_j(r_j)+\epsilon},$
where $\epsilon>0$ ensures numerical stability. The Policy optimizes the GRPO clipped surrogate using these advantages~\citep{shao2024deepseekmathpushinglimitsmathematical}. Writing $\mathbf r_t=(r_1,\ldots,r_N)$, we summarize Policy update as $\theta_{t+1}=\operatorname{GRPOUpdate} (\theta_t;x,q,Y_t,\operatorname{stopgrad}(\mathbf r_t)).$
The Generator and Judge receive no gradients from this update. Their capabilities come from Stage I, while transferring changes in shared visual and linguistic representations may help them remain coordinated with the evolving Policy. 
The Generator and Judge share momentum parameters $\phi_t$, initialized as $\phi_0=\theta_0$. After every $H$ Policy updates, the shared momentum model is updated by
\begin{equation}
    \phi_{t+1}=
    \begin{cases}
        m\phi_t+(1-m)\theta_{t+1},&(t+1)\bmod H=0,\\
        \phi_t,&\text{otherwise},
    \end{cases}
    \label{eq:momentum-transfer}
\end{equation}
where $m\in[0,1]$ is the momentum coefficient.

\textbf{Synchronization and adaptation.} Since the RL objective optimizes caption generation rather than rubric quality, we do not assume that copying Policy parameters improves either rubric role. We use a controlled transfer, inspired by MoCo~\citep{he2020momentumcontrastunsupervisedvisual}. We define the transfer rule as an exponential moving average (EMA), aiming to balance adaptation with retention of those capabilities:
\begin{equation}
    \mathcal{S}(\phi_t,\theta_{t+1})=m\phi_t+(1-m)\theta_{t+1}.
    \label{eq:ema-transfer}
\end{equation}
Here $m\in[0,1]$ is the momentum coefficient; the two terms retain the previous momentum parameters and incorporate the updated Policy parameters, respectively.

EMA retains part of the previous rubric-role state while incorporating a fraction of the updated Policy parameters. The update occurs after the current group has been scored and before a subsequent group is evaluated. Both rubric roles therefore use the same parameter version to formulate and execute each rubric. Rubric content changes with the candidate group, whereas the model parameters change only at synchronization boundaries. The cases $m=1$, $(m,H)=(0,1)$, and $m=0$ with $H>1$ correspond to frozen rubric roles, copying after every Policy update, and periodic full copying, respectively. The tracking and smoothing effects of intermediate values are analyzed below; whether the rubric roles retain their task performance requires empirical evaluation.

\subsection{Tracking--Stability Analysis}

We analyze how the momentum coefficient $m$ and synchronization interval $H$ balance tracking of the current Policy against smooth updates of the shared rubric model. 
Let $\bar{\theta}_s$ and $\bar{\phi}_s$ be Policy and momentum parameters at synchronization boundary $s$. Define the tracking error
$e_s=\lVert\bar{\theta}_s-\bar{\phi}_s\rVert$, the Policy displacement
$\Delta_s=\lVert\bar{\theta}_s-\bar{\theta}_{s-1}\rVert$, and the momentum-update magnitude
$u_s=\lVert\bar{\phi}_s-\bar{\phi}_{s-1}\rVert$.


\paragraph{Proposition 1 (Tracking--stability trade-off).}
Suppose $0\leq m<1$, each Policy update satisfies
$\lVert\theta_{t+1}-\theta_t\rVert\leq\delta$, and synchronization occurs every $H$ updates. Then
\begin{equation}
e_s
\leq
m^s e_0+
\frac{m(1-m^s)}{1-m}H\delta,
u_s
\leq
(1-m)(e_{s-1}+\Delta_s).
\label{eq:eq1}
\end{equation}
Shared initialization gives $e_0=0$, and hence
$e_s\leq mH\delta/(1-m)$. Thus, the Policy--momentum parameter gap remains bounded under bounded Policy updates, while $m$ and $H$ jointly determine how closely the rubric roles track the current Policy.

To quantify stochastic smoothing, we consider the local approximation $\bar{\theta}_s-\bar{\theta}_{s-1}=Hv+\varepsilon_s$, $\mathbb{E}[\varepsilon_s]=0$,$\operatorname{Cov}(\varepsilon_s)=H\Sigma$, 
where the mean drift $v$ is approximately constant within the local analysis window and $\varepsilon_s$ is independent across non-overlapping synchronization intervals. The covariance assumption models the aggregation of $H$ approximately independent per-step perturbations with locally stationary covariance. These stochastic assumptions are used only for the smoothing analysis and are not required for Proposition~1. 
Let $q_s=\bar{\phi}_s-\bar{\phi}_{s-1}$. In the stationary regime, $\mathbb{E}[q_s]=Hv$, $\operatorname{Cov}(q_s) = \frac{1-m}{1+m}H\Sigma.$
Compared with periodic full copying ($m=0$), EMA therefore reduces the stochastic update covariance by the factor $(1-m)/(1+m)$, while preserving the long-run mean drift. A larger $m$ produces smoother rubric-role updates but increases tracking lag, whereas a larger $H$ increases the drift and noise accumulated between synchronizations. 
Appendix~\ref{app} gives the full finite-step and stationary derivations. These results describe parameter tracking and smoothing, rather than guaranteeing that caption optimization improves rubric construction or judging.

%% file: sections/experiments.tex
\begingroup
\raggedbottom
\setlength{\textfloatsep}{12pt plus 2pt minus 2pt}
\setlength{\floatsep}{10pt plus 2pt minus 2pt}
\setlength{\intextsep}{10pt plus 2pt minus 2pt}
\makeatletter
\newcommand{\localcaptionfont}[1]{%
  \let\originalmakecaption\@makecaption
  \long\def\@makecaption##1##2{{#1\originalmakecaption{##1}{##2}}}%
}
\makeatother

\begin{table}[htbp]
\centering
\caption{Pairwise win rates (\%) against Qwen3-VL-8B across five captioning benchmarks. Average denotes the benchmark mean. Best results are bold; second-best results are underlined.}
\label{tab:pairwise-caption-comparison}
\small
\setlength{\tabcolsep}{4pt}
\resizebox{0.92\textwidth}{!}{
\begin{tabular*}{\textwidth}{@{\extracolsep{\fill}}lcccccc@{}}
\toprule
Method & PixMo-Cap & DenseFusion & CapArena & CompreCap & DOCCI & Average \\
\midrule
\multicolumn{7}{@{}l}{\textit{Captioning Baselines}} \\
\addlinespace[2pt]
ShareGPT4V-7B & 2.02 & 1.80 & 2.67 & 5.21 & 2.41 & 2.82 \\
RICO-Flash-7B & 8.89 & 5.41 & 9.00 & 9.87 & 9.04 & 8.44 \\
OmniCaptioner-8B & 10.91 & 10.62 & 12.17 & 10.05 & 11.24 & 11.00 \\
JoyCaption-8B & 11.72 & 4.81 & 12.67 & 25.85 & 14.46 & 13.90 \\
MetaCaptioner-8B & 22.63 & 23.25 & 19.00 & 32.50 & 18.47 & 23.17 \\
CapRL-InternVL-8B & 23.03 & 16.83 & 15.67 & 25.67 & 15.26 & 19.29 \\
CapRL-Qwen3VL-4B & 42.22 & 39.48 & 48.83 & 34.83 & 50.00 & 43.07 \\
Qwen3-VL-32B & 64.85 & 63.73 & 62.00 & 56.19 & 57.23 & 60.80 \\
\midrule
\multicolumn{7}{@{}l}{\textit{Rubric-based Methods}} \\
\addlinespace[2pt]
RubiCap & \underline{70.14} & \underline{67.80} & \underline{72.33} & \textbf{63.57} & \underline{75.80} & \underline{69.93} \\
EvoLM & 59.19 & 53.71 & 56.17 & 48.47 & 62.63 & 56.03 \\
\midrule
\multicolumn{7}{@{}l}{\textit{Our Framework}} \\
\addlinespace[2pt]
SFT (shared) & 55.15 & 53.31 & 51.83 & 49.19 & 54.22 & 52.74 \\
Ours & \textbf{73.94} & \textbf{70.34} & \textbf{78.00} & \underline{63.38} & \textbf{78.51} & \textbf{72.83} \\
\bottomrule
\end{tabular*}
}
\end{table}

\section{Experiments}
\label{sec:experiments}

\subsection{Experimental Setup}
\label{sec:experimental-setup}

\begin{table}[t]
\centering
\caption{Caption-based question answering across five benchmarks. Average is the mean across benchmarks. Bold and underlined values indicate the best and second-best results, respectively.}
\label{tab:caption-based-qa}
\small
\setlength{\tabcolsep}{4pt}
\resizebox{0.92\textwidth}{!}{
\begin{tabular*}{\textwidth}{@{\extracolsep{\fill}}lcccccc@{}}
\toprule
Method & CaptionQA & BLINK & TextVQA & DocVQA & ChartQA & Average \\
\midrule
Base & 80.41 & 47.05 & \underline{58.00} & 69.18 & 61.28 & 63.18 \\
RubiCap & \underline{83.28} & 49.05 & 56.85 & \underline{77.90} & \textbf{64.44} & \underline{66.31} \\
Qwen3-VL-32B & 82.92 & \textbf{50.94} & 56.69 & 75.06 & 62.16 & 65.55 \\
EvoLM & 82.00 & 49.00 & 57.08 & 77.21 & 62.44 & 65.55 \\
\midrule
SFT (shared) & 81.05 & 47.69 & 56.44 & 76.13 & 63.16 & 64.89 \\
Ours & \textbf{84.24} & \underline{50.10} & \textbf{58.33} & \textbf{79.05} & \underline{64.12} & \textbf{67.17} \\
\bottomrule
\end{tabular*}
}
\end{table}

\textbf{Implementation Details.}
We use Qwen3-VL-8B-Instruct~\citep{bai2025qwen3vltechnicalreport} as the backbone and perform full-parameter training on H100 GPUs (80GB).
Both SFT and RL datasets contain equal proportions of data from PixMo-Cap~\citep{deitke2024molmopixmoopenweights} and DenseFusion~\citep{li2024densefusion1mmergingvisionexperts}.
For SFT, we use 20,000 supervision records generated by GPT-5.5~\citep{openai2026gpt55}: 10,000 for the Policy and 5,000 each for the Generator and Judge.
We train for 300 steps with a learning rate of $1\times10^{-5}$ and a global batch size of 64.
For RL, we optimize the Policy with GRPO~\citep{shao2024deepseekmathpushinglimitsmathematical} on 5,000 samples for 600 steps at a learning rate of $1\times10^{-6}$.
We use a per-device batch size of 8 and two gradient accumulation steps.
Both stages use AdamW~\citep{loshchilov2019decoupledweightdecayregularization} with cosine learning rate decay and a 5\% warmup ratio.

\textbf{Benchmarks.}
We evaluate both caption quality and downstream task utility.
Caption quality evaluation covers PixMo-Cap~\citep{deitke2024molmopixmoopenweights}, DenseFusion~\citep{li2024densefusion1mmergingvisionexperts}, CapArena~\citep{cheng2025caparenabenchmarkinganalyzingdetailed}, CompreCap~\citep{lu2025benchmarkinglargevisionlanguagemodels}, and DOCCI~\citep{onoe2024doccidescriptionsconnectedcontrasting}, with blind ranking and ablation studies conducted on PixMo-Cap and DenseFusion.
Downstream evaluation spans caption utility (CaptionQA~\citep{yang2026captionqacaptionusefulimage}), visual perception and reasoning (BLINK~\citep{fu2024blinkmultimodallargelanguage}), and scene-text understanding (TextVQA~\citep{singh2019vqamodelsread}).
It also includes document understanding (DocVQA~\citep{mathew2021docvqadatasetvqadocument}) and chart understanding and reasoning (ChartQA~\citep{masry-etal-2022-chartqa}).
These complementary tasks test whether generated captions preserve visual information needed to answer diverse questions, complementing direct assessments of caption quality.

\textbf{Baselines.}
We compare captioning baselines with rubric-based methods. Captioning baselines include ShareGPT4V-7B~\citep{chen2024sharegpt4v}, RICO-Flash-7B~\citep{wang2025ricoimprovingaccuracycompleteness}, OmniCaptioner-8B~\citep{lu2025omnicaptionercaptionerrule}, JoyCaption-8B~\citep{fpgaminerjoycaption}, MetaCaptioner-8B~\citep{lei2026metacaptioner}, the InternVL-8B and Qwen3-VL-4B variants of CapRL~\citep{xing2026caprl}, and the larger Qwen3-VL-32B-Instruct~\citep{bai2025qwen3vltechnicalreport}. For rubric-based methods, we adapt and reproduce RubiCap~\citep{huang2026rubicap} and EvoLM~\citep{li2026evolmselfevolvinglanguagemodels} for our captioning task and training data. We select EvoLM as the alternating-training baseline because the original method shares Policy and Generator parameters, offering a closer comparison to our shared initialization. DynamicRubric~\citep{wang2026coevolvingllmevaluatorspolicies} also follows an alternating-training approach, but we do not reproduce it because its Generator training requires additional ranking annotations for anchor responses. Our RubiCap reproduction uses Gemini 3.1 Pro as the Rubric Generator and GPT-5.6-terra as the Rubric Judge. We report the shared SFT model to assess gains from subsequent reinforcement learning.

\textbf{Evaluation Metrics.}
Following CapArena~\citep{cheng2025caparenabenchmarkinganalyzingdetailed}, an LLM judge compares each method's captions against Qwen3-VL-8B-Instruct; we report win rates.
Following RubiCap~\citep{huang2026rubicap}, we blind-rank captions from five representative methods and report rank distributions and quality scores.
For question answering, we follow Prism~\citep{NEURIPS2024_cac9e747}, as adopted in CapRL~\citep{xing2026caprl}, using a fixed text-only model with captions as its sole source of visual information.
We report benchmark scores; evaluation prompts and the question-answering protocol are provided in Appendices~\ref{app:prompts} and~\ref{app:qa-protocol}.

\begin{table}[t]
\centering
\caption{Component ablation on PixMo-Cap and DenseFusion, reported as win rates (\%). Bold and underlined values indicate the best and second-best available results, respectively.}
\label{tab:component-ablation}
\small
\setlength{\tabcolsep}{3pt}
\resizebox{0.92\textwidth}{!}{
\begin{tabular*}{\textwidth}{@{\extracolsep{\fill}}ccccccc@{}}
\toprule
ID & SFT initialization & Rubrics & Rubric-role update & PixMo-Cap & DenseFusion & Average \\
\midrule
M1 & Policy only & --- & --- & 57.37 & 56.91 & 57.14 \\
M2 & Shared across roles & --- & --- & 55.15 & 53.31 & 54.23 \\
\midrule
M3 & Separate per role & Offline, fixed & No momentum update & 70.74 & 66.00 & 68.37 \\
M4 & Separate per role & Online & No momentum update & 71.31 & \underline{67.13} & \underline{69.22} \\
M5 & Separate per role & Online & EMA ($m=0.99$, $H=1$) & 65.06 & 61.80 & 63.43 \\
M6 & Shared across roles & Online & No momentum update & 68.48 & 64.73 & 66.61 \\
M7 & Shared across roles & Online & Full synchronization & \underline{71.72} & 65.73 & 68.73 \\
\midrule
M8 & Shared across roles & Online & EMA ($m=0.99$, $H=1$) & \textbf{73.94} & \textbf{70.34} & \textbf{72.14} \\
\bottomrule
\end{tabular*}
}
\end{table}

\subsection{Main Results}
\label{sec:main-results}

\begin{figure}[t]
\centering
\begin{minipage}[t]{0.67\textwidth}
\vspace{0pt}
\centering
\begin{minipage}[t][138pt][t]{\linewidth}
\includegraphics[width=\linewidth]{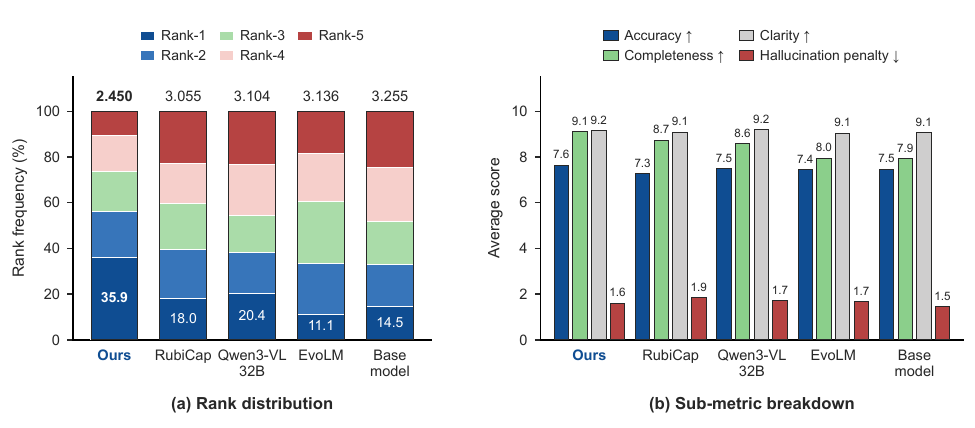}
\end{minipage}
\par\vspace{-12pt}
\setlength{\abovecaptionskip}{0pt}
\localcaptionfont{\fontsize{8.5}{10}\selectfont}
\captionof{figure}{Blind ranking on PixMo-Cap and DenseFusion: rank distributions (left) and caption quality scores (right), averaged across the two datasets.}
\label{fig:blind-ranking}
\end{minipage}\hfill
\begin{minipage}[t]{0.31\textwidth}
\vspace{0pt}
\centering
\begin{minipage}[t][138pt][c]{\linewidth}
\centering
\scriptsize
\setlength{\tabcolsep}{1.5pt}
\renewcommand{\arraystretch}{1.45}
\resizebox{\linewidth}{!}{%
\begin{tabular}{@{}llccc@{}}
\toprule
Generator & Judge & \shortstack{PixMo\\Cap} & \shortstack{Dense\\Fusion} & Average \\
\midrule
Frozen    & Frozen    & 68.48 & 64.73 & 66.61 \\
EMA       & Frozen    & 71.14 & \underline{67.00} & \underline{69.07} \\
Frozen    & EMA       & 68.27 & 63.40 & 65.84 \\
Full sync & Full sync & \underline{71.72} & 65.73 & 68.73 \\
EMA       & EMA       & \textbf{73.94} & \textbf{70.34} & \textbf{72.14} \\
\bottomrule
\end{tabular}}
\end{minipage}
\par\vspace{-12pt}
\setlength{\abovecaptionskip}{0pt}
\localcaptionfont{\fontsize{8.5}{10}\selectfont}
\captionof{table}{Role transfer; EMA $(m,H)=(0.99,1)$.}
\label{tab:role_specific_ema}
\end{minipage}
\end{figure}


\textbf{Pairwise Caption Comparison.}
Table~\ref{tab:pairwise-caption-comparison} shows that MoCo Rubric achieves a 72.83\% average win rate across five benchmarks, exceeding Qwen3-VL-32B by 12.03 percentage points with an 8B backbone. RubiCap and EvoLM also surpass the specialized captioners on average, while MoCo Rubric further exceeds them by 2.90 and 16.80 points, respectively, despite RubiCap's use of proprietary models for rubric generation and judging. EvoLM trails shared SFT on CompreCap (48.47\% versus 49.19\%), illustrating that dynamic rubric training does not improve every benchmark.
The advantage persists with GPT-5.6 Sol as the evaluator, and both LLM evaluators show high agreement with human preferences on the tested caption pairs (Appendices~\ref{app:gpt56-evaluation} and~\ref{app:human-agreement}).

\textbf{Caption-Based Visual Question Answering.}
Using captions as the sole visual input to a fixed text-only model, MoCo Rubric achieves the highest average score of 67.17 in Table~\ref{tab:caption-based-qa}, exceeding RubiCap and shared SFT by 0.86 and 2.28 points. It ranks first on CaptionQA, TextVQA, and DocVQA and second on BLINK and ChartQA, showing that the caption gains also benefit downstream question answering.


\textbf{Blind Ranking.}
In the joint evaluation of five anonymized methods, MoCo Rubric achieves the best mean rank (2.450) and highest first-place share (35.9\%) in Figure~\ref{fig:blind-ranking}. It leads in accuracy and completeness, has clarity comparable to Qwen3-VL-32B, and improves on RubiCap's mean rank (3.055) and hallucination penalty (1.859 versus 1.614).


\subsection{Analysis}
\label{sec:ablation-studies}

\textbf{Which Components Drive the Gains?}
We ablate components and combinations to assess contributions to caption quality.
In Table~\ref{tab:component-ablation}, replacing fixed offline rubrics with online rubrics (M3$\rightarrow$M4) improves average win rate by 0.85 pp, supporting criteria adapted to current Policy candidates.
Shared initialization alone offers no advantage: shared SFT underperforms Policy-only SFT (M2 vs.\ M1), and remains below separate initialization when rubric roles are frozen (M6 vs.\ M4).
Its benefit emerges with momentum transfer.
Under separate initialization, EMA reduces average win rate from 69.22\% to 63.43\% (M4$\rightarrow$M5); with shared initialization, it raises win rate from 66.61\% to 72.14\% (M6$\rightarrow$M8).
This contrast supports a common parameter foundation for transferring Policy updates to both rubric roles.
EMA outperforms full synchronization by 3.41 pp (M7$\rightarrow$M8), indicating that gradual transfer contributes to the gains.
These results highlight complementary contributions: online rubrics adapt evaluation criteria to current candidates, while shared initialization and momentum updates jointly support cross-role parameter transfer.
Training diagnostics further show fewer rollout groups with zero within-group reward variation under our method (Appendix~\ref{app:training-reward}).

\textbf{Which Roles Benefit from Momentum Transfer?}
Starting from the shared SFT checkpoint, we apply EMA to either rubric role or both. Table 4 shows that Generator-only EMA raises the average win rate from 66.61\% to 69.07\%, whereas updating only the Judge lowers it to 65.84\%. Joint EMA reaches 72.14\%, exceeding Generator-only EMA and full synchronization (68.73\%), suggesting that Judge updates help when coordinated with Generator updates through momentum transfer.

\begin{figure}[!htbp]
\centering
\begin{minipage}[c][114pt][c]{0.33\textwidth}
\centering
\includegraphics[width=0.76\linewidth]{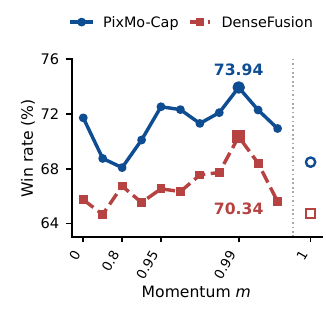}
\end{minipage}\hfill
\begin{minipage}[c][114pt][c]{0.33\textwidth}
\centering
\includegraphics[width=0.76\linewidth]{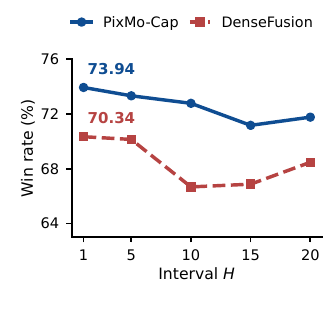}
\end{minipage}\hfill
\begin{minipage}[c][114pt][c]{0.33\textwidth}
\centering
\includegraphics[width=0.76\linewidth]{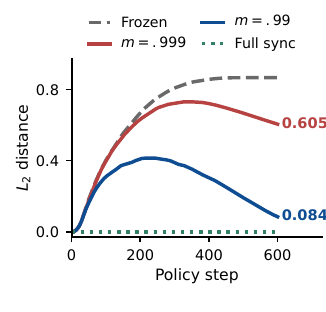}
\end{minipage}
\vspace{-20pt}
\setlength{\abovecaptionskip}{0pt}
\begin{minipage}[t]{0.9866\textwidth}
\localcaptionfont{\footnotesize}
\captionof{figure}{\textbf{Left:} Momentum coefficient analysis at $H=1$, with tested coefficients equally spaced and the frozen reference ($m=1$) shown separately. \textbf{Middle:} Synchronization interval analysis at $m=0.99$. Markers indicate measured results; lines connect adjacent tested settings. \textbf{Right:}Policy--momentum tracking at $H=1$. Frozen and full-sync curves are boundary references.}
\label{fig:momentum-analysis}
\end{minipage}\hfill
\par\vspace{8pt}
\end{figure}

\textbf{How Do Transfer Magnitude and Frequency Affect Performance?}\leavevmode\par\nopagebreak
\begin{wraptable}{r}{0.32\textwidth}
\vspace{-12pt}
\centering
\localcaptionfont{\footnotesize}
\caption{Equal-weight mean win rate (\%) across PixMo-Cap and DenseFusion in the joint $m\times H$ sweep.}
\label{tab:momentum-joint}
\scriptsize
\setlength{\tabcolsep}{1pt}
\begin{tabular*}{\linewidth}{@{\extracolsep{\fill}}cccc@{}}
\toprule
$m$ & $H=1$ & $H=5$ & $H=20$ \\
\midrule
0.95 & 69.58 & 68.18 & 69.29 \\
0.99 & \textbf{72.14} & 71.79 & 70.18 \\
0.999 & 68.28 & 66.67 & 69.48 \\
\bottomrule
\end{tabular*}
\vspace{-5pt}
\end{wraptable}
The momentum coefficient $m$ controls transfer magnitude, and the interval $H$ controls transfer frequency. Figure~\ref{fig:momentum-analysis} shows that $m=0.99$ performs best with stepwise synchronization; at this coefficient, $H=5$ remains close to $H=1$. Table~\ref{tab:momentum-joint} reports the highest mean win rate of 72.14\% at $(m,H)=(0.99,1)$, with 71.79\% at $(0.99,5)$; longer intervals reduce win rates at $m=0.99$, although the effect of $H$ is not monotonic at other coefficients. These results motivate selecting $m$ and $H$ jointly and are consistent with the tracking--stability trade-off in Section~3.4, without establishing it as the cause of the performance differences.

\textbf{How Closely Does the Momentum Model Track the Policy?}
Figure~\ref{fig:momentum-analysis} shows the parameter distance between the Policy and the momentum model at H = 1. In separate runs with m = 0.99 and m = 0.999, the distance initially increases and then levels off or declines, reaching 0.084 and 0.605, respectively, at step 600. At that step, the frozen reference computed from the m = 0.99 Policy trajectory has a distance of 0.867, while the ideal reference with full synchronization at every step has zero distance. Relative to the frozen reference on the same Policy trajectory, EMA reduces the parameter distance in later training. This observation provides empirical support for the tracking error analysis in Proposition 1 of Section 3.4.

\endgroup

%% file: sections/conclusion.tex
\section{Conclusion}
\label{sec:conclusion}

We presented a framework that coordinates caption generation, rubric construction, and judging to address cross-role mismatch in detailed image captioning.
Shared multi-task SFT establishes a foundation, while online rubrics and EMA parameter updates sustain adaptation as the Policy evolves.
Experiments show stronger overall performance than existing methods across complementary evaluations, with gains in caption quality and downstream utility.
These results support coordinated rubric adaptation as an approach to reinforcement learning for detailed image captioning.

%% file: sections/appendix_prompts.tex
\newtcolorbox{promptbox}[1]{
  breakable,
  colback=black!2,
  colframe=black!35,
  colbacktitle=black!8,
  coltitle=black,
  fonttitle=\small\bfseries,
  title={#1},
  title after break={#1 (continued)},
  boxrule=0.4pt,
  arc=1pt,
  left=6pt,right=6pt,top=5pt,bottom=5pt,
  before skip=8pt,after skip=8pt
}
\DefineVerbatimEnvironment{PromptText}{Verbatim}{
  fontsize=\footnotesize,
  breaklines=true,
  breakanywhere=true,
  breaksymbolleft={},
  breakindent=0pt
}

We provide the prompts for the Caption Policy, Rubric Generator, Rubric Judge, and two external caption evaluations.
Per-example inputs are represented by placeholders.
The Policy and Generator receive the image, denoted by \texttt{<image>} below.
Both external evaluators also receive the image, whereas the Rubric Judge receives only the caption and rubrics.

\subsubsection{Caption Generation}
\label{app:caption-prompt}

The Caption Policy generates a detailed image description using the following prompt.

\begin{promptbox}{System prompt}
\begin{PromptText}
You are an image captioning assistant. You output only plain-text image descriptions without any markdown formatting, headers, bullet points, or structural elements. Your descriptions are detailed, flowing paragraphs.
\end{PromptText}
\end{promptbox}
\begin{promptbox}{User prompt}
\begin{PromptText}
<image>Describe this image in detail.
\end{PromptText}
\end{promptbox}

\subsubsection{Rubric Generation}
\label{app:rubric-generation-prompt}

The Rubric Generator receives the image, two reference captions, and four sampled current-Policy captions.
The following template specifies the criterion, binary evaluation rule, and weight of each rubric item.

\begin{promptbox}{User prompt}
\begin{PromptText}
<image>
You generate strict, image-grounded rubrics for distinguishing the quality of image captions.

You receive one image, two strong reference captions, and four captions sampled from the current policy. The image is the final authority. The references help identify reliable details. The policy captions reveal quality differences that the rubrics should capture; none of the captions is automatically correct.

Reference caption 1:
<reference_caption_1>

Reference caption 2:
<reference_caption_2>

Current-policy caption 1:
<policy_caption_1>

Current-policy caption 2:
<policy_caption_2>

Current-policy caption 3:
<policy_caption_3>

Current-policy caption 4:
<policy_caption_4>

Goal:
Generate binary rubric items that capture important, image-verifiable differences among the policy captions. Prioritize decision boundaries that separate more accurate and precise captions from vague, incomplete, incorrect, or unsupported ones. Do not optimize for exhaustive coverage or generate easy items merely because they describe true image content.

Instructions:
1. Inspect the image and use the references to identify salient, reliable facts.
2. Compare the policy captions to identify any meaningful, image-verifiable difference that affects caption quality. Consider factual correctness, specificity, important omissions, unsupported claims, and any other distinction that materially separates better captions from worse ones.
3. Verify each difference against the image, then express the correct distinction as a self-contained rubric that also applies to future captions.
4. Preserve the most specific important fact supported by the image. Do not weaken an exact count, visible text, identity, attribute, action, or spatial relation into a broader condition simply to make more captions pass.
5. Define each Pass/Fail boundary at the highest reliable specificity supported by the image. When an important fact is precisely verifiable and that precision matters to caption quality, vague wording, an incorrect value, or omission must fail. Do not enforce uncertain, unimportant, or incidental details.
6. Choose a rubric set that reflects overall caption quality: preserve important distinctions while giving appropriate credit for accurate central content. Avoid relying entirely on error-specific checks or broad, easy-to-pass facts, and consolidate errors that arise from the same underlying fact.
7. Keep each item atomic and non-redundant. Do not create overlapping or nested criteria for the same underlying fact.
8. Judge semantic meaning rather than exact wording, except when exact visible text, numbers, names, or counts are the fact being tested.

Illustrative example:
If the image clearly shows exactly five masks and the count materially distinguishes caption quality, use one rubric requiring exactly five masks. "Multiple masks", a wrong count, or omission of the important count must fail; do not add separate broader criteria for the presence of masks. Apply this precision-preserving principle to any reliable distinction, not only counts.

Weights:
- 3.0 for a central fact or severe factual error.
- 2.0 for an important quality distinction.
- 1.0 for a useful secondary distinction.

Return only one valid JSON object inside one ```json code fence:
```json
{
  "rubrics": [
    {
      "criterion": "...",
      "evaluation_rule": "Pass if ...; fail if ...",
      "weight": 1.0
    }
  ]
}
```

Do not mention the references or policy captions in the output. Do not add extra fields or text outside the JSON code fence.
\end{PromptText}
\end{promptbox}

\subsubsection{Rubric-Based Scoring}
\label{app:rubric-judge-prompt}

The text-only Rubric Judge evaluates each caption against the supplied rubrics, returning a reason followed by a binary decision for each item.
The variables in braces are filled for each request.

\begin{promptbox}{System prompt}
\begin{PromptText}
You are a strict reward evaluator for image captions. Judge only whether the candidate caption satisfies the supplied rubrics. The rubrics are the ground truth. Do not reward unsupported claims. Do not penalize issues that are not covered by the rubrics. Give one concise, caption-grounded reason before each binary decision. Return valid JSON only.
\end{PromptText}
\end{promptbox}
\begin{promptbox}{User prompt template}
\begin{PromptText}
You are an image-caption reward judge.
Your task is to evaluate whether a generated caption satisfies each supplied rubric criterion.
Judge by semantic meaning and intent, not exact wording or keyword matching.

Generated caption:
{generated_caption}

Rubrics:
{rubrics}

Evaluation rules:
1. Evaluate every rubric independently, in the same order as listed.
2. Before pass, write a concise, specific reason grounded in the generated caption and the current rubric. For pass=1, identify how all required conditions are satisfied; for pass=0, identify the decisive unmet, contradicted, or unsupported condition.
3. Set pass=1 only if the caption satisfies both the criterion and its evaluation_rule.
4. Set pass=0 if the caption omits required information, contradicts the rule, or contains unsupported content explicitly covered by the rubric.
5. Judge only the requirements in the current rubric. Do not penalize issues that the rubric does not cover.
6. Accept synonyms, different sentence structures, and other semantically equivalent expressions.
7. Do not penalize style, fluency, grammar, or length unless the rubric explicitly requires it.

Return exactly one valid JSON object in this format:
{example_json}
The items array must contain exactly {item_count} elements, one per rubric, in the same order.
Each item must contain exactly two fields in this order: reason, then pass.
Each reason must be a non-empty JSON string containing a concise, specific, evidence-based explanation.
Each pass value must be the integer 0 or 1, not true/false and not a string.
Return valid JSON only. Do not include markdown, code fences, comments, scores, or text outside the JSON object.
\end{PromptText}
\end{promptbox}

Here \texttt{\{generated\_caption\}} is the candidate caption and \texttt{\{item\_count\}} is the number of rubric items, $N$.
The \texttt{\{rubrics\}} field lists all $N$ items in the following format; the ellipsis denotes repeated entries.
\begin{promptbox}{Rubric serialization}
\begin{PromptText}
1. weight=<weight_1>
criterion: <criterion_1>
evaluation_rule: <evaluation_rule_1>

...

N. weight=<weight_N>
criterion: <criterion_N>
evaluation_rule: <evaluation_rule_N>
\end{PromptText}
\end{promptbox}

The \texttt{\{example\_json\}} field contains exactly $N$ example items, alternating the two reason--decision patterns below.
This two-item illustration specifies the output format; it is not a judgment of a particular caption.
\begin{promptbox}{Output-format example for two rubric items}
\begin{PromptText}
{
  "items": [
    {
      "reason": "The caption states the required subject and attribute.",
      "pass": 1
    },
    {
      "reason": "The caption omits a required detail.",
      "pass": 0
    }
  ]
}
\end{PromptText}
\end{promptbox}

\subsubsection{Pairwise Evaluation for Win Rate}
\label{app:win-rate-prompt}

An external evaluator receives the image and two captions and returns \texttt{A}, \texttt{B}, or \texttt{Tie}, together with a reason.
In the stored pairwise evaluation requests, Caption A is the evaluated method's output and Caption B is the Qwen3-VL-8B-Instruct baseline output.

\begin{promptbox}{Evaluation prompt}
\begin{PromptText}
Given an image and two candidate captions, determine which caption is better.

Evaluation guidelines:
1. Precision:
   Caption should accurately match the image.
   Penalize wrong color, quantity, spatial relation, posture, etc.

2. Informativeness:
   Caption should include salient information.
   More specific descriptions are preferred when accurate.

3. Hallucination:
   Penalize descriptions of objects or elements absent from the image.

4. Attention to detail:
   Carefully inspect image details.

5. Assistive description:
   Imagine describing the image to a visually impaired person.

6. Reverse thinking:
   Ask what image the caption makes you imagine and whether it matches the actual image.

7. Ties are acceptable:
   If both captions are similarly good, output Tie.

Ignore:
- writing style or phrasing
- caption length
- grammatical variations

Caption A:
<caption_A>

Caption B:
<caption_B>

Example output:
{
  "reason": "...",
  "judgment": "A" | "B" | "Tie"
}

Return only one JSON object in exactly the same format as the example output. Do not include markdown, code fences, or any text before or after the JSON.
\end{PromptText}
\end{promptbox}

\subsubsection{Blind Ranking}
\label{app:blind-rank-prompt}

The evaluator receives the image and five captions labeled \texttt{Caption A} through \texttt{Caption E}, without model names.
The prompt below includes the four scoring dimensions, score aggregation rule, and output template used for both PixMo-Cap and DenseFusion.

\begin{promptbox}{Evaluation prompt}
\begin{PromptText}
You are an expert image captioning evaluator. Given the image above and the 5 captions below, rigorously assess each caption.

## Scoring
Score each caption on four dimensions (integers 0-10):

1. accuracy - Are the described objects, actions, text, colors, and spatial relationships factually correct for this image? Penalize for any wrong attribute, misidentified object, or incorrect action.
2. completeness - Does the caption cover all visually significant elements (main subjects, notable actions, background context, on-screen text if present)? Penalize for missing key details.
3. clarity - Is the caption well-written, specific, grammatically correct, and unambiguous? Penalize for vague language or redundancy.
4. hallucination_penalty - Does the caption assert things NOT visible in the image? 0 = zero hallucination; 10 = pervasive fabrication. Be strict: even plausible but unverifiable claims count as mild hallucination (2-4). This score is applied as a penalty.

Compute: total_score = (accuracy + completeness + clarity) / 3.0 - hallucination_penalty x 1.5

## Captions to Evaluate
Caption A:
<caption_A>

Caption B:
<caption_B>

Caption C:
<caption_C>

Caption D:
<caption_D>

Caption E:
<caption_E>

## Output Format
Respond ONLY with a single valid JSON object - no markdown fences, no extra text.
{
  "assessments": {
    "Caption A": {
      "justification": "<2-3 sentences citing specific visual evidence from the image>",
      "accuracy": "<int 0-10>",
      "completeness": "<int 0-10>",
      "clarity": "<int 0-10>",
      "hallucination_penalty": "<int 0-10>",
      "total_score": "<float>"
    },
    "Caption B": {
      "justification": "<2-3 sentences citing specific visual evidence from the image>",
      "accuracy": "<int 0-10>",
      "completeness": "<int 0-10>",
      "clarity": "<int 0-10>",
      "hallucination_penalty": "<int 0-10>",
      "total_score": "<float>"
    },
    "Caption C": {
      "justification": "<2-3 sentences citing specific visual evidence from the image>",
      "accuracy": "<int 0-10>",
      "completeness": "<int 0-10>",
      "clarity": "<int 0-10>",
      "hallucination_penalty": "<int 0-10>",
      "total_score": "<float>"
    },
    "Caption D": {
      "justification": "<2-3 sentences citing specific visual evidence from the image>",
      "accuracy": "<int 0-10>",
      "completeness": "<int 0-10>",
      "clarity": "<int 0-10>",
      "hallucination_penalty": "<int 0-10>",
      "total_score": "<float>"
    },
    "Caption E": {
      "justification": "<2-3 sentences citing specific visual evidence from the image>",
      "accuracy": "<int 0-10>",
      "completeness": "<int 0-10>",
      "clarity": "<int 0-10>",
      "hallucination_penalty": "<int 0-10>",
      "total_score": "<float>"
    }
  },
  "ranking": [
    "Caption A",
    "Caption B",
    "Caption C",
    "Caption D",
    "Caption E"
  ]
}

The "ranking" list must contain all 5 caption labels ordered from best (index 0) to worst (index 4), sorted strictly by total_score descending.
\end{PromptText}
\end{promptbox}

%% file: sections/appendix_human_agreement.tex
We assess the agreement between LLM judges and human preferences on 500 Qwen3-8B versus GPT-5.6 caption pairs, comprising 250 PixMo-Cap and 250 DenseFusion examples. Two annotators independently labeled each pair while blinded to the caption sources and to each other's judgments. Their annotations were reconciled into a final A/B/tie preference label for each pair. We measure exact agreement as the proportion of valid LLM judge responses that match these final labels; invalid responses are excluded from the denominator.

\begin{table}[htbp]
\centering
\caption{Agreement between LLM judges and final human preferences on Qwen3-8B versus GPT-5.6 caption pairs.}
\label{tab:human-preference-agreement}
\begin{tabular}{lc}
\toprule
LLM judge & Exact agreement \\
\midrule
Gemini 3.1 Pro & 429/499 (85.97\%) \\
GPT-5.6 Sol & 412/497 (82.90\%) \\
\bottomrule
\end{tabular}
\end{table}

Both judges show high agreement with the final human labels on this comparison set. These results support the use of our LLM-judge metric as a proxy for human caption preferences in this setting.

%% file: sections/appendix_case_studies.tex
\begin{center}
\includegraphics[width=0.96\linewidth]{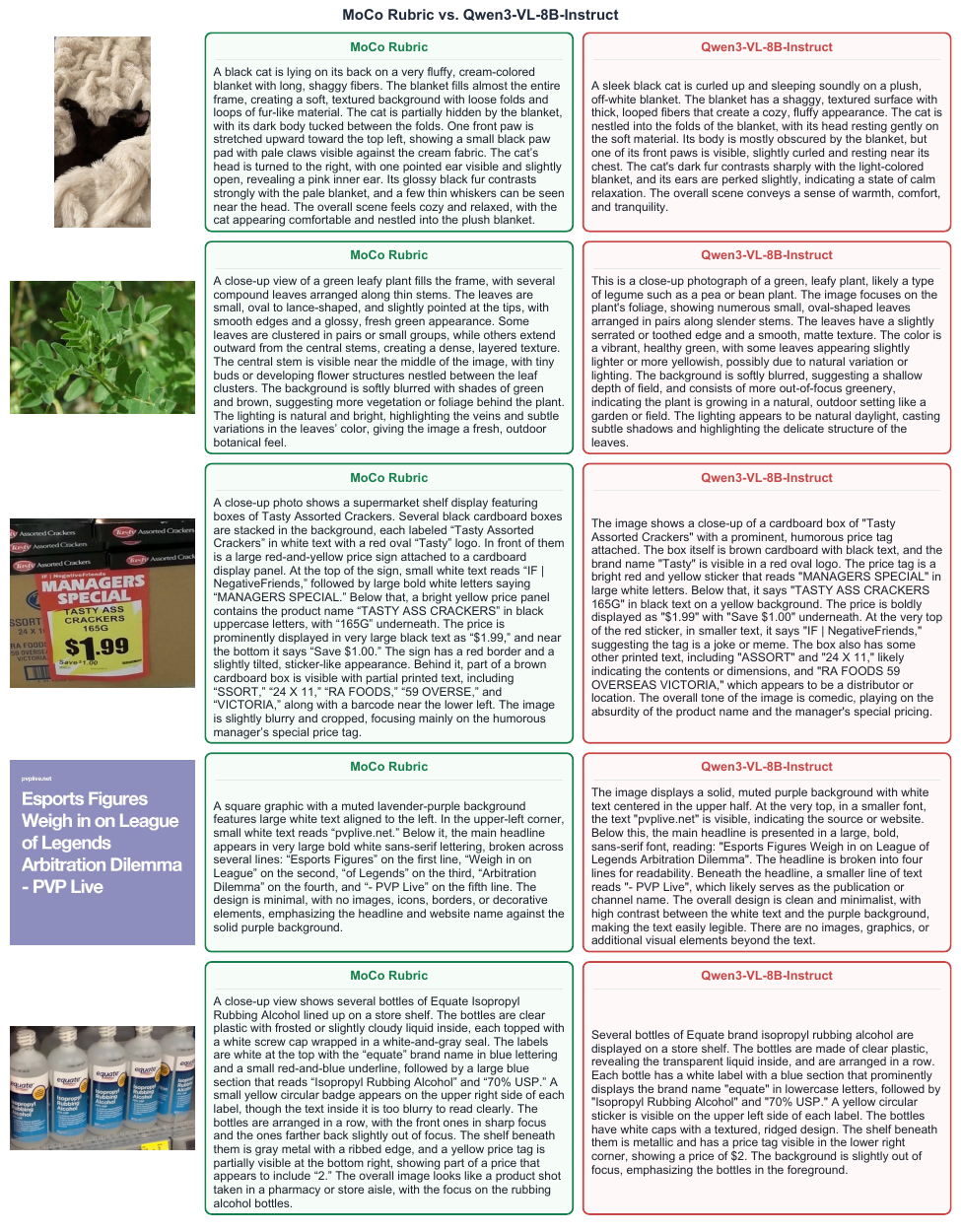}
\captionof{figure}{MoCo Rubric vs. Qwen3-VL-8B-Instruct. The selected examples illustrate differences in object pose, leaf shape, product layout, text alignment, and label position.}
\label{fig:case-study-qwen}
\end{center}

\clearpage
\begin{center}
\includegraphics[width=\linewidth]{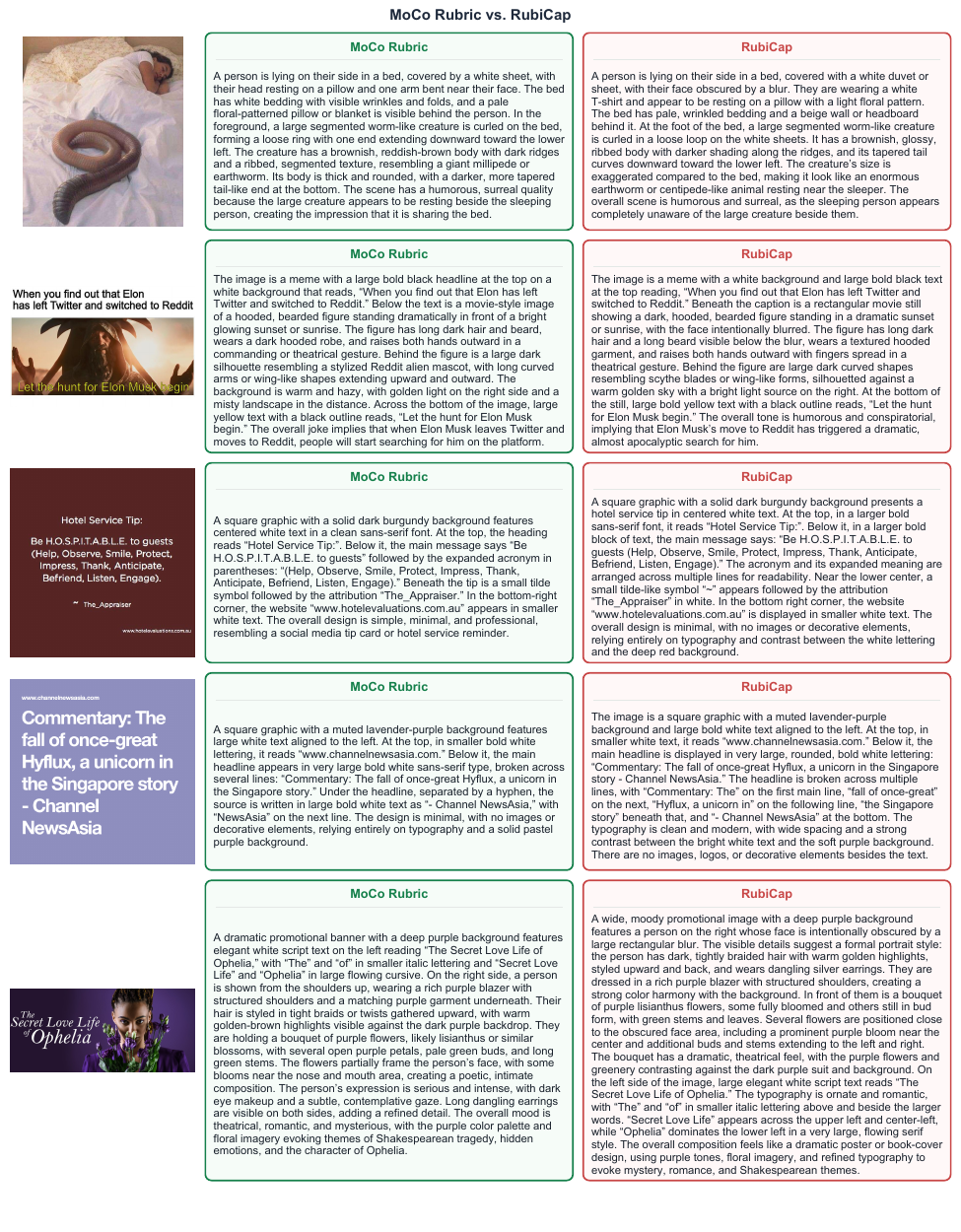}
\captionof{figure}{MoCo Rubric vs. RubiCap. The selected examples illustrate fabricated facial blurring and errors in text layout and typography.}
\label{fig:case-study-rubicap}
\end{center}

\clearpage
\begin{center}
\includegraphics[width=\linewidth]{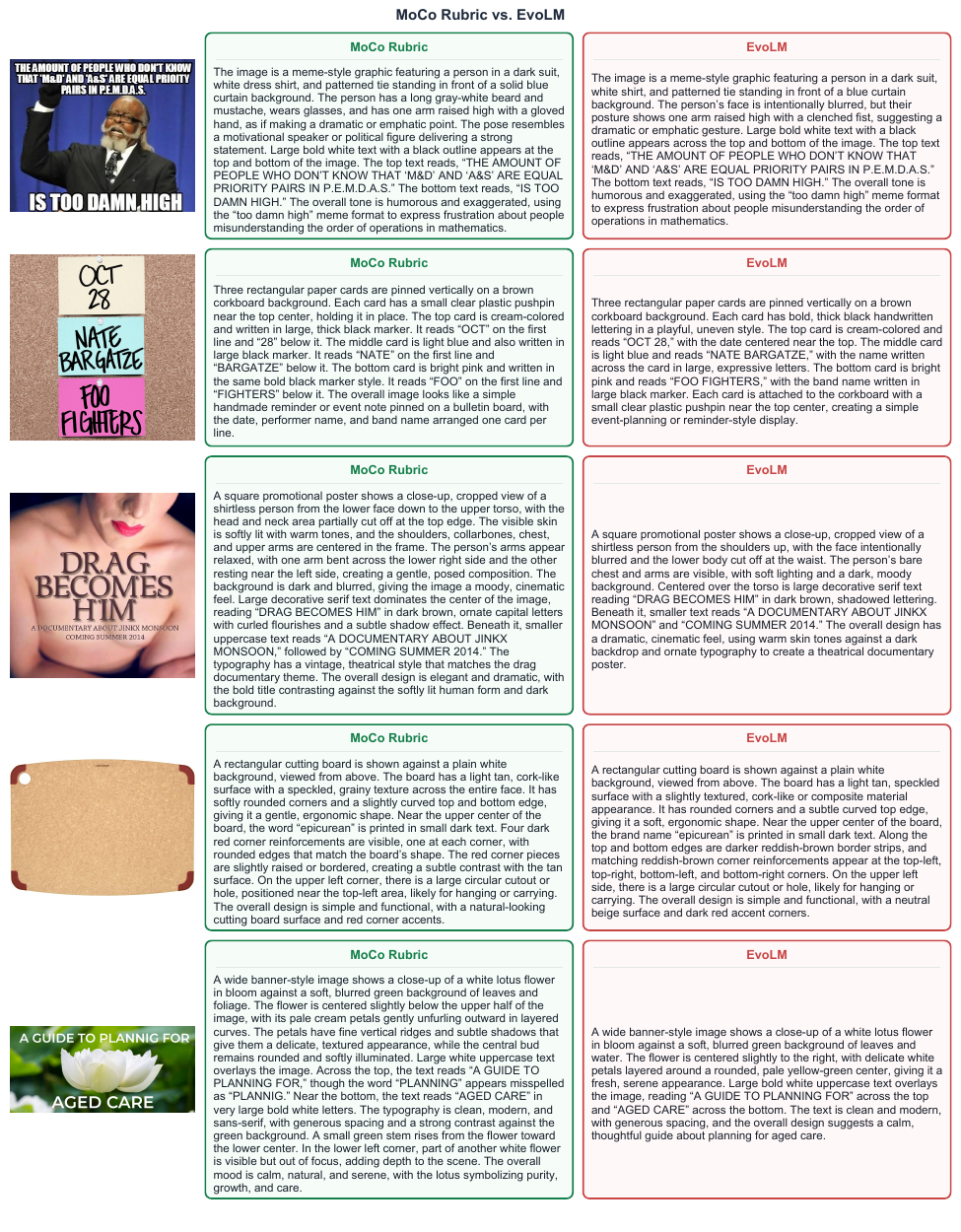}
\captionof{figure}{MoCo Rubric vs. EvoLM. The selected examples illustrate fabricated facial blurring and border details, as well as errors in text layout and transcription.}
\label{fig:case-study-evolm}
\end{center}

\clearpage